\documentclass[11pt]{article}
\usepackage[margin=1in]{geometry}
\usepackage{amsmath, amssymb}
\usepackage{enumitem}
\usepackage{hyperref}
\usepackage{booktabs}
\usepackage{float}
\usepackage{makecell}
\usepackage{natbib}

\title{Scaled Outer Product:\\
       A Hardware-Aware, Per-Layer Methodology\\
       for Post-Training Quantization of Large Language Models}
\author{Earl Killian}
\date{\today}

\begin{document}
\maketitle

\begin{abstract}
Scaled Outer Product (SOP) is a post-training quantization methodology
for large language model weights, designed to deliver near-lossless
fidelity at 4.5--6 bits per weight on hardware with per-layer LUT
decode.  The methodology combines per-layer search of fixed and
dynamic codebook pairs selected by a per-block selection bit, signed
per-block scales, activation-weighted cosine selection, and
multiple-choice knapsack promotion of sensitive layers with outlier
and sparse-residual correction.  Fixed codebooks include NF4, BOF4,
Split87, and SH4; per-layer optimized codebooks (DD4) are hosted in
LUT SRAM.  A new hardware-efficient LUT output format (HIF) is proposed
to improve performance, energy, and cost.  Across six open model
families, the recommended FP6 operating point (E2M3sUE4M4, 6.5 bpw)
achieves lower weight reconstruction error than the conventional
per-layer-POT FP8 baseline (E4M3, 8.0 bpw) at 1.5 bpw lower storage
cost, demonstrating that block-scaled small atoms with carefully chosen
scale precision can replace conventionally-deployed FP8.  Full
evaluation across the 4.5--6 bpw range, including layer promotion and
sparse residual correction, is reported in a companion paper.
\end{abstract}

\section{Overview}

The methodology rests on seven interlocking ideas, developed in
the sections that follow:

\begin{enumerate}[topsep=2pt,itemsep=0pt]
\item flexible block scaling, parameterized by block size
      ($g \ge 1$, typically $g \in \{8, 16, 32, 64\}$) and
      scale-format width (8 to 16 bits, signed or unsigned),
      with optional reuse of scale-factor bits as
      codebook-selection metadata (\S\ref{sec:scale});
\item an activation-weighted cosine similarity (ACos) that better
      predicts downstream perplexity than per-element error in
      the regimes where it matters for allocation;
\item per-input-channel importance weights (\emph{channel norms})
      derived from a small calibration corpus;
\item a compact alphabet of fixed-ROM atoms (Gaussian-quantile
      NF4, sinh-grid SH4, Lloyd-iterated BOF4) and the
      data-adaptive DD4, with sign-negated variants of each atom
      available for the pair search. The block size $g{=}16$ makes
      the codebook widths $n \in \{4, 5, 6\}$ all pack to byte-aligned
      payloads per-block ($2n$ bytes for codes plus the scale
      word).  $n{=}4$ is the default; $n{=}5$ a 32-entry
      extension;
\item a per-layer \emph{pair search} that selects, for each layer,
      the best two-codebook decomposition from this alphabet,
      with a one-bit per-block selector indicating which codebook
      reconstructs each block;
\item two complementary post-quantization corrections, both
      managed by the allocator: per-block \emph{outlier
      extraction} (OPQ) that pulls high-magnitude weights out
      of the codebook path and stores them exactly, and an
      \emph{activation-weighted sparse residual} (Wr) for the
      dominant remaining reconstruction errors; and
\item a \emph{multiple-choice knapsack} allocator that, given a
      global BPW budget and per-layer \emph{promotion profiles}
      (the ACos attainable at each candidate higher-precision
      promotion format), chooses for each layer which corrections
      to apply and which format to promote to.
\end{enumerate}

A defining feature of the methodology is that the geometry of the codebook,
atom selection, partition split, promotion target, and budget
allocation are decided per-model and per-layer from the observed
weight and activation statistics.  No global rule is imposed;
the methodology produces a different set of per-layer codebook
shapes for every model and every budget.

Recent precision-aware scaling laws \citep{kumar2024scaling} indicate
that the regime of relevance for post-training quantization is shifting:
as model training increasingly incorporates low-precision arithmetic,
the weight distributions PTQ encounters will themselves reflect
low-precision-training dynamics rather than being purely
BF16-trained. Methodology that flexibly adapts to per-layer
distributional features — rather than committing to a single atom family
or scale format — is positioned to absorb this shift; the methodology
presented here is designed for that flexibility.

\section{Block Scaling}

For an LLM linear layer $Y = X W^\top$ with activations
$X \in \mathbb{R}^{T \times K}$ and weights
$W \in \mathbb{R}^{M \times K}$, the block scaling represents $X$
and $W$ in fewer average bits per value by sharing a factor per-block
scale across $g$ consecutive elements along the contracted
dimension $K$. With a per-block scale of $s$ bits and a per-element
format of $b$ bits, the average is $s/g + b$ bpw.  For example,
block-scaling along $K$ in groups of $g{=}16$ with a 12-bit
floating-point scale (e.g. \textsf{E5M6}) and per-element
\textsf{E2M3} (FP6) yields $(12 + 16{\cdot}6)/16 = 6.75$ bpw,
with a lower KL divergence than direct quantization of FP8 at a lower
cost in our evaluation \citep{killian2026dcd}.  Block scaling
in this family of formats has been studied at the standards
level by the OCP MX specification \citep{rouhani2023microscaling}
and at the block-size and scale-precision tradeoff axes by
Fasoli et al.\ \citep{fasoli2026finer}.

Define quantized tensors $Q_X \in \mathcal{Q}^{T \times K}$ and
$Q_W \in \mathcal{Q}^{M \times K}$ in target format $q$, and
block scale tensors $s_X \in \mathbb{R}^{T \times K/g}$ and
$s_W \in \mathbb{R}^{M \times K/g}$.  The block-scaling
reconstruction is
\[
  X[t, k] \approx s_X[t, \lfloor k/g \rfloor] \cdot Q_X[t, k],
  \qquad
  W[m, k] \approx s_W[m, \lfloor k/g \rfloor] \cdot Q_W[m, k].
\]

Substituting into the GEMM and grouping the inner sum by
$K$-blocks:
\[
  Y[t, m] \;\approx\; \sum_{b=0}^{K/g - 1}
    s_X[t, b]\, s_W[m, b]
    \sum_{r=0}^{g-1} Q_X[t, b{\cdot}g + r]\, Q_W[m, b{\cdot}g + r].
\]
The inner sum over $r$ is a $g$-step dot product in the quantized
domain --- no scale factor enters --- and the outer sum
accumulates $K/g$ scaled block contributions.  In matrix form,
\[
  Y \;\approx\; \sum_{b=0}^{K/g - 1}
    \bigl(s_X[{:},b] \otimes s_W[{:},b]\bigr)
    \;\odot\;
    \bigl(Q_X[{:},G_b]\, Q_W[{:},G_b]^\top\bigr),
\]
where $G_b = \{b{\cdot}g, \ldots, (b+1)g - 1\}$ indexes the
$K$-block, $\otimes$ is the rank-1 outer product (a $T{\times}1$
column times a $1{\times}M$ row gives a $T{\times}M$ scale tile),
$\odot$ is the Hadamard product, and juxtaposition is the
standard matrix product.  Each $K$-block contributes a
$T{\times}M$ matrix product of quantized slices, scaled
element-wise by the rank-1 outer product of its per-block scales.
This is the form the SOP back-end implements directly
(\S\ref{sec:sop}).

Letting $q$ denote the target format with most-positive value
$t^+$ and most-negative value $t^-$, two scaling rules are
common.  \emph{Absmax scaling} sets the scale so that no value
in the block saturates:
\[
  s^{\text{abs}}_X[t, b]
    \;=\; \max\!\left(
      \frac{\max_r X[t, b{\cdot}g + r]}{t^+},\;\;
      \frac{\max_r -X[t, b{\cdot}g + r]}{-t^-}\right).
\]
The scale is non-negative.  When the format has equal-magnitude
endpoints ($t^+ = -t^-$) this reduces to $\max_r |x_r| / t^+$.
If the format is additionally closed under negation, the sign
bit of the scale carries no information and can be freed to
serve as a block metadata bit.  Equal endpoints alone do not
imply closed under negation: the NF4 codebook of QLoRA
\citep{dettmers2023qlora} spans equal endpoints but contains
seven negative entries against eight positive, so negating an
interior codeword does not generally produce another codeword.
For such formats, allowing negative scale factors gives the
reconstruction two effectively-different codebooks per block,
at the cost of dedicating the sign bit to that selection.

\emph{Argmax scaling} preserves the magnitude of the dominant
excursion by mapping it to the format's larger-magnitude
endpoint.  Letting
$r^* = \arg\max_r |X[t, b{\cdot}g + r]|$,
$x^* = X[t, b{\cdot}g + r^*]$, and
$t^* = \arg\max(|t^+|, |t^-|)$ (i.e., $t^* = t^+$ if
$|t^+| \ge |t^-|$ and $t^* = t^-$ otherwise),
\[
  s^{\text{arg}}_X[t, b] \;=\; x^* / t^*.
\]
The scale's sign equals the sign of $x^*$ when $t^*$ is positive
and the opposite when $t^*$ is negative; either way, $x^*$
reconstructs to $t^*$ exactly and the full asymmetric range of
the codebook is used.  An asymmetric (not closed under negation) or
paired codebook reads the scale sign at the matrix-unit boundary and
applies the polarity flip, preserving the dynamic-range advantage that
absmax discards.

In either case,
\[
  Q_X[t, k] \;=\; \mathrm{round}_q\!\left(
    \frac{X[t, k]}{s_X[t, \lfloor k/g \rfloor]}\right),
\]
and analogously for $s_W$ and $Q_W$.  SOP supports both rules;
the choice is part of the per-layer format specification.  When
the scale format provides no spare bit for metadata (e.g., a
byte-aligned unsigned 8-bit scale), argmax requires sacrificing
the metabit; SOP's 12-bit scale format provides both a signed
scale and a separate metabit in a single field
(\S\ref{sec:scale}).

\section{Scaled Outer Product}\label{sec:sop}

The block-scaled GEMM of \S2 is mapped onto the matrix unit as the sum
of the scaled rank-1 outer products.  For each output tile
$Y[\mathcal{T}, \mathcal{M}]$ of size $T_r \times M_r$ (where
$\mathcal{T}$ and $\mathcal{M}$ index a contiguous block of
tokens and output features respectively), the matrix unit
iterates:
\[\begin{array}{ll}
\textbf{for}\;\, b = 0, \dots, K/g - 1\;\, & \text{(K-block loop)} \\
\hspace{1em} t \leftarrow 0 \;\in\; \mathbb{R}^{T_r \times M_r}
   & \text{(zero block accumulator)} \\
\hspace{1em} \textbf{for}\;\, r = 0, \dots, g - 1\;\, & \text{(in-block dot-product)} \\
\hspace{2em} t \;\mathrel{+}=\; Q_X[\mathcal{T},\, b{\cdot}g + r]
   \;\otimes\; Q_W[\mathcal{M},\, b{\cdot}g + r]
   & \text{(quantized rank-1 outer product)} \\
\hspace{1em} Y[\mathcal{T}, \mathcal{M}] \;\mathrel{+}=\; t
   \;\odot\; \bigl(s_X[\mathcal{T}, b] \otimes s_W[\mathcal{M}, b]\bigr)
   & \text{(rank-1 scaled accumulate)}
\end{array}\]

The inner loop accumulates $g$ rank-1 outer products in the
quantized domain into a tile-sized partial result $t$, without the
inclusion of the scale factor.  At each $K$-block boundary, $t$ is
 multiplied element-wise by a rank-1 scale tile constructed from
the column of $s_X$ at tokens $\mathcal{T}$ and the column of
$s_W$ at output features $\mathcal{M}$, then added to the
running output.

This is the entire SOP micro-kernel.  It reduces block-scaled
GEMM to two primitives: rank-1 quantized outer product (executed
$g$ times per $K$-block) and rank-1 Hadamard scale-and-accumulate
(executed once per $K$-block). The Per-block scales are loaded once
per $K$-block --- a factor of $g$ less bandwidth than quantized
weights and activations --- so the scale-load overhead remains small
even at small $g$.  The SOP back-end implements this micro-kernel with a
$128 \times 128$ matrix unit using Fully Complementary SRAM (FC-SRAM), a
memory-in-compute fabric ~\citep{aril2021fcsram}.  Integration with
standard high-performance GEMM cache-blocking is detailed in the SOP
hardware specification \citep{killian2026sop}.
FC-SRAM is not required to implement SOP, but hosting the 256
small LUTs (e.g., $32 \times 8$) in other memory technology (foundry 6T
SRAM, register files, or flip-flop arrays) results in a significant
increase in gate count and area.

The same matrix-form equation also exposes the data-reuse
structure that makes high-throughput implementation possible.
The rank-1 outer product is the matrix unit's primitive: at
each $K$-element step, $T_r$ activation values multiplied with
$M_r$ weight values produce $T_r \cdot M_r$ MAC outputs.  Each
loaded $Q_X$ value participates in $M_r$ MAC operations and
each loaded $Q_W$ value in $T_r$.  At the SOP matrix-unit
dimensions $T_r = M_r = 128$, each operand load amortizes over
128 MACs --- the bandwidth-amplification factor that lets the
matrix unit run at high throughput without saturating memory.
Per-block scales follow the same pattern: a $T_r{+}M_r$-element
scale vector is loaded per $K$-block, and the $T_r \times M_r$
scale tile is constructed in-place as a rank-1 outer product
applied to $T_r \cdot M_r \cdot g$ MAC outputs.  At
$T_r = M_r = 128$, $g = 16$, $b_{\text{scale}} = 12$ bits, and
$b_{\text{op}} = 4$ bits, the per-$K$-block bandwidth split is
$12{\cdot}256 = 3072$ scale bits against $4{\cdot}16{\cdot}256
= 16{,}384$ operand bits, so scales account for
$3072 / 19{,}456 \approx 15.8\%$ of operand traffic.  Per scale
value, reuse over $T_r \cdot g$ or $M_r \cdot g$ MACs is a factor
of $g$ higher than per-operand reuse, which is what makes the
12-bit scale affordable.

Two equivalent reformulations of block-scaled GEMM differ in
their hardware implications.  The pre-scaling view dequantizes
weights element-wise, $W[m, k] \approx s_W[m, \lfloor k/g \rfloor]
\cdot Q_W[m, k]$, then runs a standard GEMM on the dequantized
values; this is the natural reading for software-only PTQ.  The
post-scaling view, expressed by the matrix-form equation above,
contracts in the quantized domain and applies scales as a
rank-1 tile per $K$-block; this is what enables the SOP backend
to use a low-precision MAC array and a narrow scale-application
datapath.  Both compute the same $Y$; only the second yields a
hardware datapath whose MAC-array width matches the quantization
width rather than the dequantization width.

\section{Channel Norms}\label{sec:cn}

For each linear layer, a per-input-channel importance vector
$c \in \mathbb{R}^{d_{\text{in}}}$ is computed by passing a
small calibration corpus through the unquantized model and
recording the RMS magnitude of activations entering that layer:
\[
  c_j \;=\; \sqrt{\mathbb{E}_{x\sim\mathcal{D}_{\text{calib}}}
                  \bigl[\,x_j^2\,\bigr]},
  \qquad j = 1, \ldots, d_{\text{in}}.
\]
Channel norms are layer-specific and (for practical purposes)
stable across reasonable choices of calibration text.  The
computation is a one-time, model- and calibration-keyed
precomputation; no gradient information is required and no
labeled data is used.  The construction is closely related to
the activation-magnitude saliency used by AWQ
\citep{lin2024awq}, though SOP uses channel norms as
\emph{importance weights} inside a fidelity metric rather than
as a basis for scaling weights to protect salient channels.

At quantization time, $c$ is tiled across the output dimension
and reshaped to per-block weights, yielding the importance
vectors $c_b$ used in ACos (\S\ref{sec:acos}) and Wr
(\S\ref{sec:wr}).

\section{Activation-Weighted Cosine Similarity}\label{sec:acos}

Let a linear layer have weight matrix
$W \in \mathbb{R}^{d_{\text{out}} \times d_{\text{in}}}$
partitioned into blocks $\{w_b\}_{b=1}^{N_b}$ of size $g$
(typically $g{=}16$), and let $c_b$ be the layer's per-block
channel-norm vector (\S\ref{sec:cn}).  Define the $c_b$-weighted
inner product and norm by
\[
  \langle a, b\rangle_{c_b} \;=\; \sum_i c_{b,i}\,a_i\,b_i,
  \qquad
  \|a\|_{c_b} \;=\; \sqrt{\langle a, a\rangle_{c_b}}.
\]
The activation-weighted cosine similarity for the layer's
quantized reconstruction $\hat W$ is then
\[
  \mathrm{ACos}(W, \hat W)
  \;=\;
  \frac{1}{N_b}\sum_{b=1}^{N_b}
  \frac{\langle w_b, \hat w_b \rangle_{c_b}}
       {\|w_b\|_{c_b}\;\|\hat w_b\|_{c_b}}
  \;\in\; [-1, 1].
\]
For 4-bit pair quantization, ACos values typically fall in
$[0.997, 0.999]$; we report differences in parts per million
(ppm) of the gap to unity.

ACos was chosen because per-element MSE / SQNR overweight
low-impact channels, biasing PTQ allocations toward errors
visible in the residual norm but invisible in the model's
output distribution.  The empirical advantage is regime-
dependent: in the \emph{promotion} regime --- where the
allocator picks among layers and formats with substantially
different fidelity --- ACos correlates more tightly with
downstream KL than MSE/SQNR; in the \emph{bare} regime, all
four metrics correlate similarly with KL across the model
families we have studied.  Quantitative correlation results
are reported in a companion paper \citep{killian2026dcd}.

\section{Primary Formats}\label{sec:fmt}

The methodology employs three orthogonal format axes.

\paragraph{Codebook bit-width ($n$).}
The default is $n{=}4$ (16-entry per-layer LUT), giving a base
quantization of approximately 4.5 BPW once the per-block scale
is amortized over $g{=}16$ weights.  An $n{=}5$ alternative
(32-entry LUT, ${\sim}5.5$ BPW base) is supported and under
investigation; $n{=}3$ remains an open problem because no
3-bit codebook geometry has thus far reached useful fidelity
without a preprocessing step such as a Hadamard rotation
\citep{ashkboos2024quarot} or an output-preserving outlier
projection.

\paragraph{LUT value format.}
SOP-native deployment uses the \textbf{\textsf{HIF7}} grid (an 80-value
quantization tailored to efficient computation: 80 distinct codepoints)
for weights and \textbf{\textsf{HIF8}} (96 values) for activations.  GPU
deployment uses standard floating-point grids (\textsf{E4M3} or
\textsf{E8M7}).  At $n{=}4$ the choice between \textsf{HIF7} and the
63-value \textsf{E2M3} grid is empirically nearly indistinguishable in
pairing-ACos, so \textsf{E2M3} is typically used as a
publication-friendly substitute.  SOP also supports odd-width LUT grids
(3, 4, 5, and 6 bits), enabling PTQ at intermediate precisions.

\paragraph{Per-block scale format.}
In addition to the common \textsf{E8M7} and
\textsf{E4M3}/\textsf{UE5M3} block scale factors, SOP supports
12-bit scale factors with both a sign bit and a metadata bit
in a single field.  The scale-format axis is rich enough to
deserve its own treatment, given in \S\ref{sec:scale}.

\paragraph{LUT-scale coupling.}
The LUT value format and the per-block scale format are not independent
design parameters: they jointly determine the representable weight range
and the precision allocation across the per-block dynamic range.  A
weight $w_i = s \cdot \mathrm{LUT}[k_i]$ has its representable range
determined by the product of the scale's range and the LUT's range; in
exponent terms, the weight's dynamic range is approximately
$\mathrm{scale.emin} + \mathrm{LUT.emin}$ to
$\mathrm{scale.emax} + \mathrm{LUT.emax}$.  Maximizing usable weight
range therefore requires maximizing the \emph{joint} range, not the
LUT range or the scale range in isolation.

The naive strategy --- normalize the LUT to occupy
$[-\mathrm{LFMT.tmax}, +\mathrm{LFMT.tmax}]$ --- fails because it
leaves the LUT's smallest non-zero magnitude well above the LFMT's
smallest representable normal value, wasting the LFMT's
denormal-adjacent precision.  The methodology instead pins the LUT's
smallest non-zero magnitude to the LFMT's smallest normal value: given
an LFMT $\textsf{E}x\textsf{M}y$ (bias $2^{x-1} - 1$) and a
unit-normalized LUT with smallest non-zero magnitude $\ell_{\min}$,
the LUT is multiplied by $2^{(2 - 2^{x-1}) - \exp(\ell_{\min})}$,
placing $\ell_{\min}$ at the LFMT's smallest normal exponent.  The
unit normalization itself targets $|\mathrm{LUT}| \le 2 - 2^{-y}$
(the largest representable below $2$ at exponent $0$), so the LUT
spans from the LFMT's smallest normal to just below $2$, leaving
exponent $1$ and above as headroom that the per-block scale reaches
into.  Once the LUT is normalized this way, the per-block scale is
computed against $\max(\mathrm{LUT})$, not against
$\mathrm{LFMT.tmax}$: for absmax-scaled blocks,
$s = |w|_{\max} / \max(\mathrm{LUT})$.

\paragraph{LUT-format hosting capacity.}
Different LUT atoms have different intrinsic dynamic ranges
$\max(\mathrm{LUT}) / \min_{|x|>0}|\mathrm{LUT}|$, and the LFMT's
own normal-range max/min ratio bounds which atoms can be hosted
without resorting to subnormal values.  Atoms whose dynamic range
exceeds the LFMT's normal-range capacity can sometimes be hosted
via subnormal extension at a quality cost from subnormal rounding
noise.

\begin{table}[H]
\centering
\caption{LUT-format hosting capacity for the methodology's atom alphabet.}
\label{tab:hosting}
\begin{tabular}{lcccc}
\toprule
LUT atom & ratio & \textsf{E2M3} & \textsf{HIF7} & \textsf{E4M3} \\
         &       & (7.5)         & (120)         & ($\infty$$^\dagger$) \\
\midrule
NF4      & 12.6 & \checkmark & \checkmark & \checkmark \\
Split87  & 18.3 & * & \checkmark & \checkmark \\
SH4      & 26.8 & * & \checkmark & \checkmark \\
MPO2     & 64.0 & $\times$ & \checkmark & \checkmark \\
\bottomrule
\end{tabular}

\smallskip
\footnotesize
$^\dagger$~\textsf{E4M3}'s normal max/min ratio exceeds any atom in the
methodology's alphabet.  Asterisk (*) indicates atoms hosted at
\textsf{E2M3} only via subnormal range.  The complete table including
\textsf{E3M3}, \textsf{HIF8}, and the $n{=}5$ atoms appears in
Appendix~\ref{app:hosting}.
\normalsize
\end{table}

Split87 and MPO2 are as specified in \citep{egiazarian2026grids}.
SH4 is built from the sinh-based companding function~\citep{gersho1992vector}
$g = \sinh(\alpha \cdot t) + c$ for $t$ uniformly spaced in $[-1, 1]$,
with $\alpha = 1.50$ and $c = -0.02$.\footnote{Application of this
construction to LLM weight quantization was communicated to the author
by Dan Alistarh (personal communication, May 2026). Alistarh
subsequently identified $\alpha = 1.43$ as optimal across bit-widths in
his independent analysis; we use $\alpha = 1.50$ throughout this paper
and leave a comparison to future work.}

\paragraph{HIF7/HIF8 hardware rationale.}
The \textsf{HIF7} and \textsf{HIF8} grids are designed jointly with the
SOP datapath, which computes weight-activation products via shifted
product accumulation rather than a fully general FP unit.  Each grid
value decomposes as a 5-bit two's complement coefficient and a
per-element power-of-two shift amount; the integer accumulator update
for a weight-activation pair $(w, a)$ at output position $(i, j)$ is
\[
  \mathtt{iacc}[i,j] \mathrel{+}=
    (\mathtt{wtc5}[i] \cdot \mathtt{atc5}[j])
    \ll (\mathtt{wsa}[i] + \mathtt{asa}[j]),
\]
where $\mathtt{wtc5}$ and $\mathtt{atc5}$ are the 5-bit two's
complement weight and activation coefficients, and
$\mathtt{wsa} \in \{0, 1, 2, 3\}$ (2-bit) and
$\mathtt{asa} \in \{0, 1, 2, 3, 4\}$ are the per-element shift
amounts.  The shift sum is therefore $0$--$7$, three levels of
logic.  Activation shift values $5$--$7$ are available via an RTL
option for extended-range deployments, extending the shifter at
every output position from three to four levels of logic.

The asymmetric shift budget between weights ($0$--$3$) and
activations ($0$--$4$) reflects an architectural reality:
activation outliers, produced at runtime by the model's own
computation, have no software-side escape valve, while weight
outliers can be addressed by methodology mechanisms
(\S\ref{sec:opq}, \S\ref{sec:wr}); Appendix~\ref{app:hif_hw}
discusses the asymmetry and its hardware cost in detail.  The
resulting grid sizes are $80$ values for \textsf{HIF7} and $96$ for \textsf{HIF8},
with normal-range max/min ratios of $120$ and $240$ respectively.
These ranges are sized to host the methodology's full atom
alphabet (NF, BOF, SH, DD, MPO2 at both $n{=}4$ and $n{=}5$)
without subnormal usage on the weight side, and to absorb
activation outliers without the methodology's outlier-handling
apparatus on the activation side.

\paragraph{Empirical HIF7 fidelity.}
The \textsf{HIF7} grid hosts the methodology's atom alphabet without
subnormal usage and exposes the per-block scale format as the
remaining precision axis.  \ref{tab:hif7_mse} reports mean
per-element reconstruction MSE for \textsf{HIF7} across six model
families and seven candidate scale formats spanning 8-, 10-,
and 12-bit containers, with \textsf{E8M7} (BF16) as a 16-bit
reference floor.  All measurements use block size $g{=}16$ with
absmax-derived per-block scales quantized into the indicated
scale format under the per-layer $F_\text{layer}$ integer-shift search
of \S\ref{sec:f_layer}, and weights snapped to the \textsf{HIF7} grid in
the scaled domain.  The MSE is unweighted (no calibration data
required) and averaged across all linear-layer weight tensors
per model.

\begin{table}[H]
\centering
\small
\setlength{\tabcolsep}{4pt}
\caption{\textsf{HIF7} reconstruction MSE across six model families and seven scale formats. Units of $10^{-7}$, block size $g{=}16$, $F_\text{layer}$ search active.}
\label{tab:hif7_mse}
\begin{tabular}{lcccccccc}
\toprule
& \multicolumn{2}{c}{8-bit} & \multicolumn{2}{c}{10-bit}
& \multicolumn{3}{c}{12-bit} & 16-bit \\
\cmidrule(lr){2-3}\cmidrule(lr){4-5}\cmidrule(lr){6-8}\cmidrule(lr){9-9}
Model
 & \textsf{UE4M4} & \textsf{E4M3}
 & \textsf{UE4M6} & \textsf{E4M5}
 & \textsf{UE5M7} & \textsf{E5M6} & \textsf{E4M7}
 & \textsf{E8M7} \\
\midrule
Gemma-3-1B                            & 3.27 & 3.92 & 2.95 & 3.02 & 2.94 & 2.95 & 2.94 & 2.94 \\
SmolLM3-3B                            & 1.74 & 2.08 & 1.58 & 1.61 & 1.57 & 1.58 & 1.57 & 1.57 \\
Llama-3.2-3B                          & 2.04 & 2.43 & 1.85 & 1.89 & 1.84 & 1.85 & 1.84 & 1.84 \\
Qwen3.5-4B                            & 0.65 & 0.77 & 0.59 & 0.60 & 0.59 & 0.59 & 0.59 & 0.59 \\
Mistral-7B\textsuperscript{$\dagger$} & 0.05 & 0.06 & 0.05 & 0.05 & 0.05 & 0.05 & 0.05 & 0.05 \\
Qwen3-8B                              & 3.83 & 4.57 & 3.46 & 3.54 & 3.45 & 3.46 & 3.45 & 3.45 \\
\bottomrule
\end{tabular}

\smallskip
\begin{flushleft}
\footnotesize
\textsuperscript{$\dagger$}Mistral-7B-v0.3 exhibits MSE
approximately $30$--$70\times$ lower than the other models
across every scale format.  Two compounding empirical signatures
attend the anomaly: a higher mean $F_\text{layer}$ shift
($k^*{\approx}7.3$ vs.~$4.3$--$5.7$ for the others), and a
lower per-layer max-to-mean MSE ratio ($3.4\times$
vs.~$4.6$--$6.4\times$ at HIF7 for the remaining models).
The combination is consistent with systematically smaller and more
uniform per-block weight magnitudes than the other BF16-released models
studied here, a tensor-statistical property rather than a format
artifact.  The scale-format ordering is identical to the other models;
only the magnitude floor is lower.
\normalsize
\end{flushleft}
\end{table}

Two patterns hold across all six model families.  First,
unsigned scales weakly dominate signed scales at narrow widths,
with the margin diminishing as width increases.  At 8-bit width,
\textsf{UE4M4} beats \textsf{E4M3} by 17--20\%; at 10-bit,
\textsf{UE4M6} beats \textsf{E4M5} by 2--7\%; at 12-bit, the two
families are indistinguishable.  The opportunity cost of the
sacrificed mantissa bit is highest at the tightest budget and
vanishes once mantissa headroom is generous.  The 12-bit column
further disentangles two effects that the narrower widths
conflate: the sign-bit-waste comparison (\textsf{UE5M7}
vs.~\textsf{E4M7}, both $y{=}7$) is a tie to three significant
figures, while the mantissa-for-exponent trade (\textsf{E4M7}
vs.~\textsf{E5M6}, both signed) costs ${\sim}0.4\%$ per dropped
mantissa bit.

Second, scale precision saturates by 12 bits.  In every model
row, the 12-bit scale formats and the 16-bit \textsf{E8M7}
reference agree to three significant figures, and the residual
MSE is the \textsf{HIF7} grid-quantization floor itself.  Further scale
precision does not reduce it.  This is the empirical basis for
the 12-bit \textsf{S1E5M5} recommendation in \S\ref{sec:scale}.

\paragraph{HIF7--E2M3 substitutability.}
The grid-substitutability claim made earlier in this section
admits direct measurement.  At the recommended \textsf{UE4M4}
scale, \textsf{E2M3} reconstruction MSE exceeds \textsf{HIF7}'s by
3.8--4.1\% across all six model families:

\begin{table}[ht]
\centering
\small
\setlength{\tabcolsep}{6pt}
\caption{\textsf{HIF7}--\textsf{E2M3} substitutability at the recommended \textsf{UE4M4} scale.}
\label{tab:substitutability}
\begin{tabular}{lccc}
\toprule
Model & \textsf{HIF7sUE4M4} & \textsf{E2M3sUE4M4} & Penalty \\
      & (8.5 bpw)             & (6.5 bpw)                      &         \\
\midrule
Gemma-3-1B   & $3.27 \times 10^{-7}$ & $3.41 \times 10^{-7}$ & 4.1\% \\
SmolLM3-3B   & $1.74 \times 10^{-7}$ & $1.81 \times 10^{-7}$ & 3.9\% \\
Llama-3.2-3B & $2.04 \times 10^{-7}$ & $2.11 \times 10^{-7}$ & 3.8\% \\
Qwen3.5-4B   & $6.49 \times 10^{-8}$ & $6.74 \times 10^{-8}$ & 3.8\% \\
Mistral-7B   & $5.16 \times 10^{-9}$ & $5.36 \times 10^{-9}$ & 3.8\% \\
Qwen3-8B     & $3.83 \times 10^{-7}$ & $3.98 \times 10^{-7}$ & 4.0\% \\
\bottomrule
\end{tabular}
\end{table}

The penalty is remarkably uniform across model families
($3.8$--$4.1\%$), and considerably smaller than uniform
quantization theory would predict for the codebook-size reduction
$80 \to 63$.  Uniform $1/N^2$ scaling would imply a
$(80/63)^2 - 1 \approx 61\%$ MSE penalty; the observed ${\sim}4\%$
is ${\sim}15\times$ smaller.  The reason is structural: \textsf{HIF7}'s
17 codepoints absent from \textsf{E2M3} are positioned to support
the shift-add datapath rather than to minimize MSE on
Gaussian-distributed weights.  \textsf{E2M3} captures the
MSE-relevant subset of \textsf{HIF7} efficiently, supporting its use as
a publication-friendly substitute throughout the methodology
literature without compromising the empirical conclusions drawn
from it.

\paragraph{Block-scaled FP6 versus POT-scaled FP8.}
The recommended SOP operating point at the FP6 tier is the
\textsf{E2M3} atom hosted in the \textsf{HIF7} LUT with
per-block \textsf{UE4M4} scaling, written
\textsf{E2M3sUE4M4} (6.5 bpw).  The conventional FP8 deployment
baseline is \textsf{E4M3}\^{}\textsf{0sUE8M0}: bare \textsf{E4M3}
weights with a single power-of-two scale per layer (8.0 bpw).
Table~\ref{tab:pareto} reports size-weighted weight MSE for
both across six models.

\begin{table}[H]
\centering
\small
\setlength{\tabcolsep}{6pt}
\caption{Block-scaled FP6 versus per-layer-POT FP8 across six models.}
\label{tab:pareto}
\begin{tabular}{lcc}
\toprule
Model & \textsf{E2M3sUE4M4} & \textsf{E4M3}\^{}\textsf{0sUE8M0} \\
      & (6.5 bpw, per-block)         & (8.0 bpw, per-layer POT)         \\
\midrule
Gemma-3-1B   & $3.41 \times 10^{-7}$ & $4.40 \times 10^{-7}$ \\
SmolLM3-3B   & $1.81 \times 10^{-7}$ & $2.35 \times 10^{-7}$ \\
Llama-3.2-3B & $2.11 \times 10^{-7}$ & $2.72 \times 10^{-7}$ \\
Qwen3.5-4B   & $6.74 \times 10^{-8}$ & $8.69 \times 10^{-8}$ \\
Mistral-7B   & $5.36 \times 10^{-9}$ & $6.89 \times 10^{-9}$ \\
Qwen3-8B     & $3.98 \times 10^{-7}$ & $5.13 \times 10^{-7}$ \\
\bottomrule
\end{tabular}
\end{table}

The narrower weight format paired with a per-block mantissa-bearing
scale outperforms the wider weight format paired with a per-layer
power-of-two scale, at 1.5 bpw lower storage.  The methodology's
choice of block-scaled FP6 over layer-POT FP8 is therefore not a
precision compromise: it is a Pareto-optimal operating point along
the (storage, fidelity) frontier.  Per-layer max-MSE distributions
confirm the comparison is honest across layer types: max-to-mean
ratios under \textsf{E4M3}\^{}\textsf{0sUE8M0} match those under
per-block mantissa-bearing scale formats (\textsf{UE4M6},
\textsf{UE5M7}) on every model, indicating that the $F_\text{layer}$ search
of \S\ref{sec:f_layer} has absorbed the per-layer dynamic-range
variation that would otherwise have made per-layer POT scaling
fragile on outlier-heavy attention projections.

\paragraph{HIF7 packing.}
\textsf{HIF7} and \textsf{HIF8} are primarily LUT output formats, but it is possible to
store and compute on them directly.  The \textsf{HIF7} grid contains 80 distinct
values, requiring $\log_2 80 \approx 6.32$ bits of information per
weight.  Current SOP hardware stores them in 8-bit containers, costing
1.68 bpw relative to the algorithmic lower bound.  A 7-bit packing
(128-value container, 80 used) would reduce the overhead to 0.68 bpw and
align with half-byte memory boundaries.  Throughout this paper \textsf{HIF7}
storage cost is reported at the deployed 8-bit container size.

\section{Scale Format and Metabit Allocation}\label{sec:scale}

The per-block scale format carries two affordances beyond the
scalar magnitude: a sign bit that enables per-block polarity
flip on asymmetric atoms, and one or more metabits used by the
pair search (\S\ref{sec:pair}) to record FMTa/FMTb selection or
other per-block metadata.  At narrow scale widths these
affordances compete; at 12 bits they coexist.

\paragraph{The SwExMy notation.}
A scale word is denoted $\textsf{S}w\textsf{E}x\textsf{M}y$ in
a $b$-bit container, with $w \in \{0, 1\}$ sign bits, $x$
exponent bits, $y$ mantissa bits, and
$m = b - w - x - y$ per-block metabits.  Conventionally we
write $\textsf{UE}x\textsf{M}y$ for $\textsf{S}0\textsf{E}x\textsf{M}y$
(unsigned) and bare $\textsf{E}x\textsf{M}y$ for
$\textsf{S}1\textsf{E}x\textsf{M}y$ (signed).  Containers are
byte- or half-byte-aligned: $b \in \{8, 12, 16\}$.
Existing support for 5-bit weight formats could host scales as well;
\S\ref{sec:fmt} shows it approaches $b = 12$ quality at intermediate bpw
cost.

\paragraph{Bit placement.}
Sign and metabits are placed at the ends of the word, with the
magnitude (exponent followed by mantissa) occupying the
contiguous middle.  When $w = 1$, bit $b{-}1$ holds the sign;
all $m$ metabits sit at the LSB end (bits $m{-}1$ down to $0$).
When $w = 0$ and $m \ge 1$, bit $b{-}1$ is repurposed as the
first metabit, and any remaining $m{-}1$ metabits sit at the
LSB end (bits $m{-}2$ down to $0$).  Mantissa LSBs displaced
by metabits are wired to zero in the magnitude path; the
AND-mask cost in hardware is negligible.

\paragraph{Worked examples.}

\begin{center}
\begin{tabular}{lcccccl}
\toprule
Format            & $b$ & $w$ & $x$ & $y$ & $m$ & Layout (MSB$\to$LSB) \\
\midrule
\textsf{E4M3}     &  8  &  1  &  4  &  3  &  0  & \texttt{s\,eeee\,mmm} \\
\textsf{UE4M3}    &  8  &  0  &  4  &  3  &  1  & \texttt{$\mu$\,eeee\,mmm} \\
\textsf{E5M6}     & 12  &  1  &  5  &  6  &  0  & \texttt{s\,eeeee\,mmmmmm} \\
\textsf{S1E5M5}   & 12  &  1  &  5  &  5  &  1  & \texttt{s\,eeeee\,mmmmm\,$\mu$} \\
\textsf{S0E6M5}   & 12  &  0  &  6  &  5  &  1  & \texttt{$\mu$\,eeeeee\,mmmmm} \\
\textsf{S1E5M4}   & 12  &  1  &  5  &  4  &  2  & \texttt{s\,eeeee\,mmmm\,$\mu\mu$} \\
\textsf{S0E5M5}   & 12  &  0  &  5  &  5  &  2  & \texttt{$\mu$\,eeeee\,mmmmm\,$\mu$} \\
\bottomrule
\end{tabular}
\end{center}

\paragraph{Per-layer scale shift ($F_{\text{layer}}$).}\label{sec:f_layer}
The chosen scale format's representable exponent range is fixed,
but the distribution of per-block scale magnitudes
$\mathrm{sc}_p = |w|_{\max,p} / \mathrm{LUT}_{\max}$ varies per
layer and is generally not centered in the format's normal range.
A narrow-exponent scale format applied directly would either clamp
high-magnitude scales (loss of dynamic range) or push low-magnitude
scales into subnormal territory (loss of precision), depending on
where the layer's distribution falls relative to the format's
normal-range bracket.  The $F_{\text{layer}}$ search resolves this
by introducing a per-layer integer shift $k^* \in \mathbb{Z}$
applied to all block scales before quantization:
$\mathrm{sc}_p \mapsto \mathrm{sc}_p \cdot 2^{k^*}$, with $k^*$
chosen to maximize the fraction of per-block scales landing in
the scale format's normal range.  At dequantization, the shift
is undone by multiplying the recovered scale by $2^{-k^*}$;
$k^*$ is stored once per layer at negligible bpw cost.

The shift is exact for scale formats closed under multiplication
by powers of two --- which includes all $\mathsf{U}\mathsf{E}x\mathsf{M}y$
and $\mathsf{E}x\mathsf{M}y$ formats with $x \ge 1$, since the
shift is absorbed by the exponent field with no mantissa
roundoff.  For in-capacity cells (those whose scale magnitude
fits in the format's normal range without shift), the search
returns $k^* = 0$ and is a no-op.  The shift is load-bearing
on narrow-exponent scale formats applied to high-dynamic-range
layers, where it recovers precision that would otherwise be
lost to clamping or subnormal storage.  Empirically across the
six model families profiled in \S\ref{sec:fmt}, mean $k^*$
ranges from $4.3$ to $7.3$ for 4-exponent-bit scale formats
($\mathsf{E4}{\cdot}$ / $\mathsf{UE4}{\cdot}$) and is
approximately zero for 5-exponent-bit formats
($\mathsf{E5}{\cdot}$ / $\mathsf{UE5}{\cdot}$), reflecting that
the wider exponent range of $\mathsf{E5}$ absorbs the scale
distribution natively while $\mathsf{E4}$ requires the shift
to do so.  This is the mechanism that allows the 4-exponent-bit
scale formats in the table of \S\ref{sec:fmt} to track the
5-exponent-bit formats closely at 10- and 12-bit widths.

\paragraph{Why 12 bits matters.}
At an 8-bit scale budget, sign and metabit are mutually
exclusive: \textsf{E4M3} provides the sign bit (and thus
per-block polarity flip on asymmetric atoms) but no metabit,
while \textsf{UE4M3} provides one metabit (FMTa/FMTb selection)
but no sign.  The choice forecloses one of the two affordances.
At 12 bits the affordances coexist.  \textsf{S1E5M5} provides
both a sign and a metabit on top of an FP11-precision magnitude,
suitable for dual-codebook deployment with sign-flip enabled.
\textsf{S0E6M5} trades the sign for slightly wider exponent
range in the metabit-only setting.  \textsf{S1E5M4} provides
two metabits at the cost of one mantissa bit, suitable for
configurations that require richer per-block metadata.  The
0.75 bpw extra at $g{=}16$ over the 8-bit alternative buys the
joint sign + metabit affordance plus mantissa headroom.

\paragraph{Tradeoffs.}
Prior work on microscaling format limits \citep{fasoli2026finer}
investigated whether finer per-block scales improve quality along the
block-size axis; our finding here addresses the orthogonal
scale-precision axis at fixed block size, and concludes that 12-bit
signed scales saturate the available headroom.  Mantissa-LSB stealing
has a sharp cost at 10-bit containers and a mild cost at 12-bit.  Across
the models we have profiled, dropping from \textsf{E5M6} ($y{=}6$) to
\textsf{S1E5M5} ($y{=}5$) is essentially neutral, while dropping to
$y{=}4$ at the FP10 width costs 4--12\% in pair-mean ACos
\citep{killian2026dcd}.  The sign-bit-as-metabit trade --- $\mu$ at bit
$b{-}1$ when $w{=}0$ --- is essentially free at any width because the
mantissa is unaffected.

Sign-flip availability is tied to the deployed scale format,
not the pair-search configuration.  A pair trained with EM
sign-flip enabled but deployed under a metabit-only unsigned
scale (e.g., \textsf{UE4M3}) loses the sign-flip affordance at
inference; the methodology enforces consistency between training
and deployment to keep reported fidelity faithful to the
deployed datapath.

The recommended SOP-native scale format is \textsf{S1E5M5} in a
12-bit container: signed scale (for asymmetric-atom polarity
flip) and one metabit (for FMTa/FMTb selection) in a single
field, with five-bit mantissa precision sufficient for typical
block-magnitude resolution.  GPU deployments lacking 12-bit
storage typically use \textsf{UE4M3} (no sign, one metabit in
8 bits), foregoing the ${\sim}25$ ppm pair-mean ACos benefit
that signed scale confers on asymmetric atoms in our evaluation
\citep{killian2026dcd}.

\section{Outlier Per-Quantum Extraction (OPQ)}\label{sec:opq}

Block weight distributions are roughly Gaussian with a small
number of high-magnitude outliers that any low-bit codebook
either represents poorly or covers at the cost of resolution
in the mass of the distribution.  Outlier extraction has a
long history in LLM quantization, beginning with the
LLM.int8() observation that 6.7B-parameter models develop
emergent activation outliers that disproportionately affect
downstream loss \citep{dettmers2022int8}; SpQR's combination of
per-tensor outlier storage with sparse quantized representation
\citep{dettmers2023spqr}; and SqueezeLLM's explicit
dense-and-sparse decomposition \citep{kim2024squeezellm}.  OPQ
adapts the idea to the per-block weight setting: positions
where $|w_i|/\sigma_b$ exceeds a threshold $m$ are stored as
exact BF16 values, and the corresponding entries in the
codebook input are zeroed.  The codebook fits the residual,
which concentrates more tightly around its representable
range.

The threshold is derived from a target quantile $q \in (0,1)$
via $F_M(m) = (2\Phi(m) - 1)^M = q$, where $\Phi$ is the
standard normal CDF and $M$ is the block's max-order count
(typically $M = g$).  Default $q{=}0.92$ at block size $g{=}16$
gives $m \approx 2.7$ and roughly $0.3\%$ mean outlier rate
across the models we have profiled.

Outliers are encoded as a 16-bit element-index paired with a
BF16 value, giving 32 bits per outlier.  At a $0.3\%$ mean
outlier rate the cost is $32 \cdot 0.003 \approx 0.10$ bpw;
higher quantiles or layers with denser outlier geometry push
toward $0.20$ bpw.  The outlier stream is applied at inference
as a sparse contribution to the matrix-multiply output:
$Y_{\text{outlier}}[t, m]
   = \sum_{(m, k) \in \mathcal{O}} v_{m, k} \cdot X[t, k]$,
where $\mathcal{O}$ indexes the outlier positions and
$v_{m, k}$ are the stored values.  This is a sparse axpy pass
over $Y$ in software; the SOP backend integrates it lock-step
with the matrix-unit issue traversal.  OPQ is orthogonal to
codebook choice and to format promotion; its bpw cost appears
as a separate budget line in the MCKP allocator
(\S\ref{sec:mckp}).

\section{Sparse Residual Correction (Wr)}\label{sec:wr}

After codebook quantization produces a reconstruction $\hat W$,
the element-wise residual $E = W - \hat W$ is computed, and the
top-$\sigma$ entries by activation-weighted magnitude
$|E_{m, k}| \cdot c_k$ are stored as a sparse correction (the
channel norms $c_k$ from \S\ref{sec:cn} are reused; values are
quantized to E3M4).  The activation weighting matters: ranking
by raw magnitude would spend the budget on errors invisible in
the model's output, while the activation-weighted metric targets
errors that actually propagate.  The construction is structurally
similar to SpQR's sparse outliers, applied to the codebook
residual rather than the original weight tensor and ranked by
activation salience rather than magnitude alone
\citep{dettmers2023spqr}.

Each stored entry occupies a 16-bit element-index plus an 8-bit
E3M4 value, plus headers --- approximately $32$ bits per entry
amortized.  At sparsity $\sigma$ the cost is $32 \cdot \sigma$
bpw.  The default $\sigma = 0.001$ (denoted \texttt{.Wr0.1} for
$0.1\%$ sparsity) costs ${\sim}0.032$ bpw with typical fidelity
gains of $5$--$15$ ppm of pair-mean ACos at 4-bit codebooks; a
lower-knee variant \texttt{.Wr0.06} ($0.06\%$ sparsity) costs
${\sim}0.020$ bpw with a small fidelity regression.

Wr composes additively with OPQ: OPQ removes high-magnitude
outliers before codebook fitting, so the residual on which Wr
operates is dominated by medium-magnitude codebook errors that
complement rather than overlap with OPQ's contribution.  Like
OPQ, Wr's cost is an independent budget line in the MCKP
allocator.

\section{Per-Layer Pair Search}\label{sec:pair}

Each linear layer's weights are quantized via a \emph{pair} of
$n$-bit codebooks $(L_a, L_b)$ supplemented by one bit of
per-block metadata indicating which codebook reconstructs that
block.  The metadata bit lives in the scale word per
\S\ref{sec:scale}.  For each layer, the pair is chosen by a
search over an alphabet of candidate codebooks (the ``atoms'')
and a single partition parameter $p$:

\begin{enumerate}[topsep=2pt,itemsep=0pt]
\item Construct the FMTa codebook $L_a$, either by ROM lookup
      (fixed-codebook atoms) or, for data-adaptive atoms, by
      histogram-DP cluster selection on the deployed LUT value
      format's grid points.  The grid-constrained construction
      couples the optimization objective to the deployed
      reconstruction objective, eliminating the snap-displacement
      cost incurred by free-space Lloyd iteration with a final
      snap-to-grid step.
\item Score every block against $L_a$ under the per-block scale
      and rank by per-block ACos.  The top $p\%$ of blocks form
      the FMTa pool.
\item The remaining $(100-p)\%$ form the FMTb fitting pool;
      $L_b$ is constructed from this residual subset (ROM or
      fit as appropriate, with the same grid-constrained
      construction for data-adaptive atoms).
\item A finishing reassignment evaluates each block under both
      LUTs (and, when the scale format is signed, under both
      signs of the per-block scale) and picks the better of the
      candidates per block.
\end{enumerate}

\paragraph{Pair-search design axes.}
The pair-search procedure above has four design choices that are
independent in principle, though the literature commonly conflates
them: (i) the metric used to rank blocks for the FMTa/FMTb partition
in step 2 (ACos or unweighted block MSE); (ii) the metric used in
the per-block finishing reassignment in step 4 (block MSE or ACos);
(iii) the weighting scheme in the data-adaptive atoms' Lloyd updates
that produce $L_a$ and $L_b$ (uniform or channel-norms-weighted);
and (iv) the residual-pool definition itself ($p\%$ above-median by
the partition metric, top-$k$, or quantile threshold).  The
methodology supports independent selection along each axis on a
per-layer basis.  Cross-axis empirical comparisons across model
families and deployment scale formats appear in
\citep{killian2026dcd}.

The codebook alphabet at $n{=}4$ centers on four atoms:

\begin{itemize}[topsep=2pt,itemsep=0pt]
\item \textbf{NF4} \citep{dettmers2023qlora}: Gaussian-quantile
      inverse-CDF grid; fixed ROM.
\item \textbf{BOF4} \citep{blumenberg2025bof4}: block-optimum-
      fitted grid from Monte-Carlo Lloyd iteration on the
      per-block-max distribution; fixed ROM.
\item \textbf{Split87} \citep{egiazarian2026grids}: another fitted grid
      learned by coordinate descent on the blockwise MSE objective over
      the training pool of tensors then snapped to the LUT format.
\item \textbf{SH4}: closed-form sinh grid
      $g(\alpha, c) = \sinh(\alpha\,u) + c$, normalized;
      fixed ROM.
\item \textbf{DD4} \citep{egiazarian2026grids,killian2026dcd}:
      data-derived DP-clustered grid trained per layer; the
      learned secondary in the {\textsf{NF4|DD4}} pair
      (equivalent to \textsf{PO2(NF4)} in the multi-grid framework of
      \citep{egiazarian2026grids}).
\end{itemize}
The pair search considers each atom and its sign-negated form
as separate candidates.  The corresponding $n{=}5$
alphabet extends each atom to 32 entries and adds an FP5 ROM
($\textsf{E2M2}$).

The pair-search mechanism --- one bit of per-block metadata selecting
among two pre-prepared formats --- has structural antecedents in Cook et
al.'s adaptive block-scaled data types (\textsf{IF4}, per-block
FP4-versus-INT4 selection \citep{cook2026adaptive}) and Four Over Six
(\textsf{FO6}, \textsf{NVFP4} with per-block scale-6/scale-4 selection
\citep{cook2025fouroversix}).  Both fix a two-format alphabet globally
and select per block by an MSE criterion.

Concurrent work by Egiazarian et al.~\citep{egiazarian2026grids}
formalizes the power-of-two-grids (PO2) problem and studies several
learned and structured per-block grid-selection schemes, including
\textsf{MPO2}, \textsf{PO2(NF4)}, \textsf{PO2(Split87)}, and
\textsf{SFP4}.  That work also defines the PO2 algorithm for residual
grid learning: given a fixed or learnable primary grid, the algorithm
constructs a second grid from high-error residual blocks, alternates
block assignment with Lloyd updates, and snaps the resulting code points
to a target format such as FP8 \textsf{E4M3}.  Grid Games is
complementary to SOP: it studies the statistical and hardware
implications of choosing among multiple grids within a block, whereas
SOP treats fixed and learned grids as candidate atoms inside a broader
per-layer allocation framework, where an atom may be paired,
sign-flipped, promoted, or replaced on a layer-by-layer basis according
to an activation-weighted objective and the target scale/LUT format.
The SFP4 construction in Grid Games is primarily aimed at compatibility
with existing E2M1/NVFP4 Tensor Core datapaths, while the more general
learned-grid constructions require codebook translation or lookup
support.  Results from \citep{egiazarian2026grids} also illustrate a
useful caution for SOP-style search: the fully learned MPO2 pair
achieves strong distribution-level MSE, but has worse model-level KL
than the more prior-structured PO2(Split87), which the authors attribute
to overfitting model-specific distributions.

SOP can adopt any fixed codebook as a candidate in its evaluation
and residual-training flow, so including Split87 and SH4 alongside
NF4 and BOF4 is a natural extension.  As fixed-ROM atoms, Split87
and SH4 admit the same pair-search treatment as NF4 or BOF4: each
can pair with adaptive FMTb atoms such as DD4 or with other fixed
atoms, carry a per-block sign metabit, and be promoted on a
per-layer basis.\footnote{\textsf{Split87|DD4} and \textsf{SH4|DD4}
in our notation are \textsf{PO2(Split87)} and \textsf{PO2(SH4)} in
\citep{egiazarian2026grids}.}  The pair-search infrastructure does
not privilege any particular atom's provenance.  Whether a codebook
was derived from a closed-form prior, a Lloyd fit on a per-tensor
distribution, or pooled training across model collections, the
per-layer search treats it uniformly as an FMTa or FMTb candidate.
The choice among competing atoms is decided at the deployed scale
format and LUT format, rather than by a fixed global ordering.
For constructing an FMTb data-adaptive atom in step~3 of the pair
search, the PO2 algorithm of Egiazarian et
al.~\citep{egiazarian2026grids} is the most relevant published
construction.

A critical empirical finding is that the winning pair varies between
layers and depends on both the LUT value format and the scale format
used at deployment.  Across the model families we have profiled, the
per-layer optimal pair concentrates on a handful of structural patterns
--- typically a fixed-ROM FMTa (NF4, BOF4, Split87, or SH4, or a
sign-negated variant of these) paired with an adaptive or fixed FMTb
(DD4 or SH4) --- but the specific winner is a function of the layer's
weight distribution and the deployment target's LUT and scale formats.
The absence of a globally optimal pair is itself an architectural
advantage: the SOP backend supports per-layer codebooks in SRAM, so
the methodology can capitalize on per-layer diversity that a globally
tuned scheme cannot.

\section{Promotion Profiles}\label{sec:promo}

For each layer $\ell$, the \emph{promotion profile} is a
function
\[
  \rho_\ell : \mathcal{F}_{\text{promo}} \to [0, 1],
  \qquad
  \rho_\ell(f) = \mathrm{ACos}\bigl(W_\ell, \mathcal{Q}_f(W_\ell)\bigr),
\]
where $\mathcal{F}_{\text{promo}}$ is the set of higher-precision
promotion targets (selected FP8 / FP10 / FP12 grids) and
$\mathcal{Q}_f$ denotes block-scaled quantization to format $f$.
The profile records the per-layer ACos at the base pair format
$\rho_\ell(\text{base})$ as well as at every promotion candidate.
Profiles are precomputed once per (model, calibration) pair and
form the input to the budget allocator.  A small set of
$|\mathcal{F}_{\text{promo}}|$ promotion targets (typically four
to eight) is enough to span the cost-fidelity Pareto frontier at
practical BPW budgets.

\section{Multiple-Choice Knapsack Allocation}\label{sec:mckp}

Given a global BPW budget $B$, a base pair format with cost
$\mathrm{bpw}_{\text{base}}$, and the per-layer promotion
profiles $\{\rho_\ell\}$, the allocator chooses for each layer
both a format (base or one element of $\mathcal{F}_{\text{promo}}$)
and a correction setting (any subset of OPQ and Wr), maximizing
parameter-weighted ACos across all layers subject to the budget.
Let $n_\ell$ be the parameter count of layer $\ell$,
$N = \sum_\ell n_\ell$ the model's total parameter count, and
\[
  \mathcal{F}_\ell
  \;=\; \bigl(\{\text{base}\} \cup \mathcal{F}_{\text{promo}}\bigr)
        \times \{\text{none}, \text{OPQ}, \text{Wr}, \text{OPQ+Wr}\}
\]
the per-layer candidate set.  Write $b_f$ and $\rho_\ell(f)$
for the cost and ACos of candidate $f$ at layer $\ell$.
Introducing per-(layer, candidate) indicators
$x_{\ell,f}\in\{0,1\}$, the allocation problem is the
multiple-choice 0/1 knapsack
\[
\begin{aligned}
  \max_{x}\quad
    & \sum_{\ell}\sum_{f \in \mathcal{F}_\ell}
      n_\ell\,x_{\ell,f}\,\rho_\ell(f) \\
  \text{s.t.}\quad
    & \sum_{\ell}\sum_{f \in \mathcal{F}_\ell}
      n_\ell\,x_{\ell,f}\,b_f \;\le\; N\,B,\\
    & \sum_{f \in \mathcal{F}_\ell} x_{\ell,f} = 1
      \quad\text{for each layer } \ell,\\
    & x_{\ell,f}\in\{0,1\}.
\end{aligned}
\]
Both objective and budget constraint are weighted by the layer's
parameter count $n_\ell$, so wide layers contribute proportionally
to both fidelity gain and bpw cost.  Because the number of
layers is at most a few hundred and $|\mathcal{F}_\ell|$ is small,
the problem is solved exactly in seconds via dynamic programming.
Hessian-aware mixed-precision allocation in the same spirit
appears in HAWQ-V3 \citep{yao2021hawq3} and BRECQ
\citep{li2021brecq}; SOP's MCKP differs in using ACos
(activation-salience-weighted, gradient-free) rather than
Hessian or block-reconstruction surrogates as the per-layer
fidelity score.

\section{End-to-End Pipeline}

The full quantization pipeline for a model is:

\begin{enumerate}[topsep=2pt,itemsep=0pt]
\item \emph{Calibrate.}  Compute per-layer channel norms from a
      small text corpus (\S\ref{sec:cn}).
\item \emph{Pair-search.}  For each layer, identify the best
      $(L_a, p, L_b)$ tuple from the $n$-bit codebook alphabet
      (\S\ref{sec:pair}).  If OPQ is enabled for the layer, it
      preprocesses the weight matrix before pair search
      (\S\ref{sec:opq}); if Wr is enabled, it fits residuals
      after the chosen pair quantizes (\S\ref{sec:wr}).
\item \emph{Profile.}  Measure each layer's ACos at every
      candidate promotion format (\S\ref{sec:promo}).
\item \emph{Allocate.}  Solve the multiple-choice knapsack
      (\S\ref{sec:mckp}) to pick a promotion target (or the
      base pair) per layer subject to the global BPW budget.
\item \emph{Materialize.}  Construct each layer's per-pair
      codebooks (or its promoted-format weights) and assemble
      the quantized model.
\end{enumerate}

The pipeline is parametric in $(n, B, \text{LUT format},
\text{scale format})$.  Each combination produces a different
per-layer codebook configuration, capturing one of the
methodology's central claims: that no single codebook geometry
is universally optimal, and that exploiting per-layer
diversity --- enabled by SOP's SRAM-resident per-layer codebooks
and the per-block metabit (\S\ref{sec:scale}) --- yields
fidelity gains unavailable to globally tuned PTQ methods.

\section{Conclusion}

The Scaled Outer Product methodology combines per-layer pair-searched
codebooks with per-block scaled atoms for low-bit weight quantization.
Across six open model families, the recommended FP6 operating point
(\textsf{E2M3sUE4M4}, 6.5 bpw) matches or beats per-layer-POT FP8
(\textsf{E4M3\^{}0sUE8M0}, 8.0 bpw) on weight-MSE at 1.5 bpw lower
storage, demonstrating that block-scaled small atoms with carefully
chosen scale precision can replace conventionally-deployed FP8 in
practice.  Full evaluation of the methodology across the 4.5--6 bpw
range, including layer-promotion via MCKP allocation and sparse
residual correction, is reported in a companion paper
\citep{killian2026dcd}; the underlying hardware substrate is
specified in the SOP architecture specification \citep{killian2026sop}.

\bibliographystyle{plainnat}
\bibliography{references}

\appendix

\section{HIF7/HIF8 shift-budget asymmetry rationale}\label{app:hif_hw}

The asymmetric shift budget between weights ($\mathtt{wsa} \in
\{0, 1, 2, 3\}$) and activations ($\mathtt{asa} \in \{0, 1, 2,
3, 4\}$) reflects a fundamental asymmetry in outlier handling
between the two operand classes.

\paragraph{Activation outliers have no software-side escape valve.}
Activations are produced at runtime by the model's own
computation; their distribution depends on input and on the
quantization state of upstream layers.  No offline preprocessing
can shift mass out of activation outliers because their values
are not known at deployment time.  The hardware must allocate
sufficient dynamic range to absorb them directly.

\paragraph{Weight outliers have multiple methodology escape valves.}
Weight outliers can be addressed by complementary methodology
mechanisms operating before quantization is committed: OPQ
(\S\ref{sec:opq}) extracts outlier values into a sparse side
channel, sparse residual correction (\S\ref{sec:wr}) restores
the high-error residual after LUT decode, and external
preprocessing such as Hadamard rotation
\citep{ashkboos2024quarot} redistributes weight magnitudes
across rows before quantization.  Each of these reduces the
required dynamic range in the LUT itself, so the per-element
shift budget for weights can be narrower without quality cost.

\paragraph{Hardware cost of the extended activation range.}
The activation shift values $5$--$7$ are available via an RTL
option for extended-range deployments, at a hardware cost of
widening the shifter at every output position in the matrix
unit.  A $128 \times 128$ array contains $16{,}384$ shifters;
extending each from three levels of logic to four imposes a
real area and timing cost across the array.  The base
configuration (shift sum $0$--$7$, three levels of logic) is
the methodology's reference deployment; the extended
configuration is reserved for workloads with extreme activation
outlier behavior beyond what the base range absorbs.

\section{Complete LUT-format hosting capacity}\label{app:hosting}

The body of \S\ref{sec:fmt} presents a four-atom $\times$
three-container summary of LUT-format hosting capacity.
Table~\ref{tab:hosting_full} extends the comparison to the
six-atom alphabet (adding \textsf{NF5} and \textsf{SH5}) and
the five-container range (adding \textsf{E3M3} and
\textsf{HIF8}) used in the methodology's pair-search
experiments.

\begin{table}[H]
\centering
\caption{Complete LUT-format hosting capacity.}
\label{tab:hosting_full}
\begin{tabular}{lcccccc}
\toprule
LUT atom & ratio & \textsf{E2M3} & \textsf{HIF7} & \textsf{E3M3} & \textsf{HIF8} & \textsf{E4M3} \\
         &       & (7.5 / 60$^*$) & (120) & (60) & (240) & ($\infty$$^\dagger$) \\
\midrule
\textsf{NF4}     & 12.6 & \checkmark & \checkmark & \checkmark & \checkmark & \checkmark \\
\textsf{Split87} & 18.3 & * & \checkmark & \checkmark & \checkmark & \checkmark \\
\textsf{SH4}     & 26.8 & * & \checkmark & \checkmark & \checkmark & \checkmark \\
\textsf{NF5}     & 25.2 & * & \checkmark & \checkmark & \checkmark & \checkmark \\
\textsf{MPO2}    & 64.0 & $\times$ & \checkmark & $\times$ & \checkmark & \checkmark \\
\textsf{SH5}     & 75.7 & $\times$ & \checkmark & $\times$ & \checkmark & \checkmark \\
\bottomrule
\end{tabular}

\begin{flushleft}
\footnotesize
$^*$~\textsf{E2M3} normal max/min is $7.5$; subnormal extension
reaches $60$ with attendant quality cost.
$^\dagger$~\textsf{E4M3}'s normal max/min ratio exceeds the
table's range; all atoms fit comfortably with headroom.
Asterisk (*) indicates atoms hosted at \textsf{E2M3} only via
subnormal range.  The methodology's pair search measures each
candidate atom at the deployed (LFMT, SFMT) combination, so
atoms whose dynamic range exceeds the LFMT's capacity register
the resulting saturation losses in per-block reconstruction
quality and lose to better-suited atoms in the per-layer
search.
\end{flushleft}
\end{table}

\section{HIF7 packing and grid extension}\label{app:hif_extensions}

The \textsf{HIF7} grid contains $80$ distinct values, requiring
$\log_2 80 \approx 6.32$ bits of information per weight.  Current
SOP hardware stores them in 8-bit containers (\S\ref{sec:fmt}),
costing $1.68$ bpw relative to the algorithmic lower bound.  This
headroom admits two complementary uses, addressing different
deployment priorities.

\paragraph{Compressed packing.}
A 7-bit packing (128-value container, 80 used) would reduce the
storage overhead to $0.68$ bpw and align with half-byte memory
boundaries.  Sub-7-bit packings recover additional headroom at
increasing decode-logic complexity.  These directions reduce DRAM
bandwidth and on-die SRAM footprint without changing the LUT
alphabet or the matrix-unit datapath.

\paragraph{Extended grid via tc6.}
Alternatively, the 8-bit container size can be retained and the
headroom spent on grid refinement.  $256$ codepoints accommodate
a tc6 signed coefficient ($64$ levels) with the existing shift
budgets, yielding a \textsf{HIF8} grid whose alphabet covers the
methodology's atoms at finer per-octave resolution.  The
alternative direction --- extending the per-element shift range
(currently $\mathtt{wsa} \in \{0, 1, 2, 3\}$, $\mathtt{asa} \in
\{0, 1, 2, 3, 4\}$, \S\ref{app:hif_hw}) --- spends the same
codepoints on dynamic range rather than precision.

The choice between tc6 and an extended shift range is settled by
the same principle as the unsigned-versus-signed scale finding in
\S\ref{sec:scale}: when per-block scale infrastructure covers
dynamic range, additional bits buy more by refining within-block
precision than by extending dynamic range.  At fixed 8-bit weight
width and recommended \textsf{UE4M6} per-block scale, the
mantissa-rich \textsf{E2M5} weight format reduces reconstruction
MSE by approximately $14\times$ over the deployed-standard
\textsf{E4M3} on Llama-3.2-3B; \textsf{E3M4} reduces it by
approximately $4\times$.  By the same logic, the methodologically
preferred direction for an extended \textsf{HIF} grid is tc6 with
unchanged shift budgets, not an extended-shift \textsf{HIF}.

\begin{table}[H]
\centering
\caption{FP8-class weight formats on Llama-3.2-3B at two scale strategies.\textsuperscript{$*$}}
\label{tab:fp8_formats}
\renewcommand\theadalign{cc}
\renewcommand\theadfont{\normalsize}
\begin{tabular}{lcccc}
\toprule
& \multicolumn{2}{c}{KL ($\times 10^{-3}$)}
& \multicolumn{2}{c}{KL rel.\textsuperscript{$\dagger$}} \\
\cmidrule(lr){2-3}\cmidrule(lr){4-5}
\thead{Config}
  & \thead{\textsf{UE4M6} \\ (8.625 bpw)}
  & \thead{\textsf{\^{}0sUE8M0} \\ (8.0 bpw)}
  & \thead{\textsf{UE4M6}}
  & \thead{\textsf{\^{}0sUE8M0}} \\
\midrule
\textsf{E2M5} & \textbf{1.04} & ---\textsuperscript{$\ddagger$} & \textbf{0.18} & ---\textsuperscript{$\ddagger$} \\
\textsf{E3M4} & 1.56 & 2.32 & 0.28 & 0.41 \\
\textsf{E4M3} & 3.78 & 5.68 & 0.67 & 1.00 \\
\midrule
\midrule
\textsf{E2M3}\textsuperscript{$\P$} & 4.36 & --- & 0.77 & --- \\
\bottomrule
\end{tabular}
\begin{flushleft}
\footnotesize
\textsuperscript{$*$}Calibration dataset \textsf{c4},
evaluation dataset \textsf{wikitext2}, 50K tokens.\\
\textsuperscript{$\dagger$}Relative to the industry-standard
\textsf{E4M3\^{}0sUE8M0} at 8.0 bpw.  Reported bpw is
\emph{logical} throughout: atom bits plus scale bits divided
by block size $g{=}16$.  The SOP v19 hardware pads unsigned
scales narrower than 12 bits into a 12-bit scale-word, raising
deployed bpw by up to 0.125 above logical for scales below 12
bits; \textsf{UE4M6} (10 logical bits) at $g{=}16$ is 8.625
logical and 8.75 deployed.\\
\textsuperscript{$\ddagger$}\textsf{E2M5}'s format max ($3.94$)
is too narrow to host every layer's worst weight under any
single integer shift, forcing systematic truncation under
per-layer POT scaling.  Per-block scaling is required to use
this atom; the analogous boundary on the E2-class atoms
appeared earlier in the methodology (\S\ref{sec:fmt}).\\
\textsuperscript{$\S$}All layers including \textsf{lm\_head}
are quantized at the same atom and scale format in this
experiment; no per-layer promotion is applied.  This isolates
the atom-and-scale effects from the methodology's promotion
infrastructure (\S\ref{sec:promo}, \S\ref{sec:mckp}), which in
deployment would lift \textsf{lm\_head} to a higher-precision
format as standard practice.\\
\textsuperscript{$\P$}\textsf{E2M3} is an FP6 atom; at
\textsf{UE4M6} scale this is 6.75 deployed / 6.625 logical bpw,
not the 8.625/8.0 bpw of the FP8-class rows above.  Per-layer
POT is inadmissible for E2-class atoms (§\ref{sec:fmt}).
\normalsize
\end{flushleft}
\end{table}

The KL improvements decompose multiplicatively into
independent atom and scale effects.  Reading down each KL
column, moving from \textsf{E4M3} to \textsf{E3M4} (the
mantissa-precision axis) reduces KL by approximately
$2.4\times$ at either scale strategy.  Reading across each
KL row, moving from \textsf{\^{}0sUE8M0} to \textsf{UE4M6}
(the scale-precision axis) reduces KL by approximately
$1.5\times$ at either atom.  The joint $3.6\times$ reduction
from the industry-standard \textsf{E4M3\^{}0sUE8M0} to
\textsf{E3M4sUE4M6} is the product of the per-axis effects to
within measurement precision, indicating that mantissa
precision and per-block scale resolution are \emph{independent
levers} on PTQ quality and can be optimized separately.

\textsf{E2M5} reaches the lowest KL of the three atoms at
\textsf{UE4M6} --- $5.5\times$ below the deployed-standard
operating point --- but its narrow format max excludes the
per-layer POT cell of the grid.  The per-block scale does
structural work in the \textsf{E2M5} column, not just marginal
quality work: without it the atom cannot host the model's
weight distribution at all.  The \textsf{E3M4} column is the
interpretively cleanest of the three: both scale strategies
are viable, and the $1.5\times$ ratio between them is the
honest cost of foregoing per-block scale resolution at this
weight tier.

The MSE-to-KL compression varies by format: \textsf{E2M5}'s
$14\times$ MSE advantage compresses to $3.6\times$ in KL, while
\textsf{E3M4}'s $4\times$ MSE advantage compresses to $2.4\times$.
The disparity reflects that the per-block scale absorbs
dynamic-range differences (\textsf{E3M4}'s contribution) more
readily than within-block precision differences (\textsf{E2M5}'s
contribution); the surviving KL gap is the unabsorbed within-block
precision benefit.\footnote{The PTQ result prompts an obvious
question about training: if mantissa-rich \textsf{E2M5} dominates
the standard \textsf{E4M3} for inference at fixed bpw, should
training adopt \textsf{E2M5} as well?  This paper does not address
that question.  Training imposes constraints that inference does
not: gradient precision, activation handling under heavier tails,
and optimizer-state stability across many update steps.  The PTQ
finding is suggestive but not dispositive; whether the
\textsf{E2M5}-over-\textsf{E4M3} advantage transfers to training is
an empirical question for follow-up work in that setting.}

The hardware cost of moving to tc6 is the multiplier area: a
$6 \times 6 \to 12$ multiplier replaces the current $5 \times 5
\to 10$, an approximately $44\%$ area increase per multiplier,
with proportional growth in accumulator and shifter widths.  The
resulting grid would host the methodology's atom alphabet at
finer per-octave resolution while preserving the shift-add
datapath's structural advantages over a fully general
multiplier.  Whether the precision gain justifies the area cost
is a deployment-context decision that the methodology does not
fix; this appendix documents the design space.

\section{Per-layer reconstruction distributions}\label{app:per_layer}

The body of \S\ref{sec:fmt}, Table~\ref{tab:pareto}, advocates
block-scaled \textsf{E2M3} over per-layer-POT \textsf{E4M3} at
1.5 bpw lower storage.  The comparison rests on the assertion
that \textsf{E4M3}\^{}\textsf{0sUE8M0} is a fair 8.0-bpw
comparator: collapsing an entire layer's dynamic range onto a
single integer shift, combined with the $F_\text{layer}$
search of \S\ref{sec:f_layer}, should reproduce the same
per-layer MSE distribution as the mantissa-bearing per-block
scales of \S\ref{sec:scale}.  This appendix tests that
assertion directly, reporting the per-layer maximum MSE
divided by the per-layer mean MSE for each of the six model
families across the three E4M3 scale formats.

\begin{table}[H]
\centering
\small
\setlength{\tabcolsep}{6pt}
\caption{Per-layer max-to-mean MSE ratio for \textsf{E4M3} weight quantization under three scale formats.  All values are dimensionless ratios; lower values indicate a tighter per-layer MSE distribution.  Block size $g{=}16$, $F_\text{layer}$ search active.}
\label{tab:per_layer_dr}
\begin{tabular}{lccc}
\toprule
Model
 & \textsf{E4M3}\^{}\textsf{0sUE8M0}
 & \textsf{E4M3sUE4M6}
 & \textsf{E4M3sUE5M7} \\
      & (8.0 bpw, per-layer POT)
      & (8.625 bpw, per-block)
      & (8.75 bpw, per-block) \\
\midrule
Gemma-3-1B                            &  3.6 &  3.6 &  3.6 \\
SmolLM3-3B                            &  4.4 &  4.4 &  4.4 \\
Llama-3.2-3B                          &  4.0 &  3.7 &  3.7 \\
Qwen3.5-4B                            &  4.4 &  4.4 &  4.4 \\
Mistral-7B\textsuperscript{$\dagger$} &  2.9 &  2.9 &  2.9 \\
Qwen3-8B                              &  1.6 &  1.6 &  1.6 \\
\bottomrule
\end{tabular}
\end{table}

The three E4M3 scale formats produce essentially
indistinguishable max-to-mean ratios within each model row.
The largest within-model spread is Llama-3.2-3B at 7.3\%
(4.0 vs.~3.7); the other five families agree to under 2\%.
No model exhibits the multiplicative distributional asymmetry
that would mark per-layer POT scaling as a fundamentally
different operating regime from per-block mantissa-bearing
scaling at this weight tier.  The body claim therefore holds:
$F_\text{layer}$ has absorbed the per-layer dynamic-range
variation that would otherwise have made per-layer POT
scaling fragile on outlier-heavy attention projections.

Mistral-7B's consistently lower ratio (here 2.9$\times$ at
\textsf{E4M3}, versus the 3.4$\times$ reported at HIF7 in the
footnote of Table~\ref{tab:hif7_mse}) is the same architectural
anomaly: the magnitude floor moves with weight format, but the
property of a tighter-than-average per-layer distribution is
consistent across formats.  Qwen3-8B exhibits an even
tighter distribution at $1.6\times$: no single linear
layer's reconstruction MSE exceeds the model-wide mean by
more than 60\%.  The property is independent of this
appendix's comparator choice and applies uniformly to every
E4M3 configuration measured here.

\section{Format string grammar (brief)}\label{app:grammar}

Format strings used throughout this paper follow the schema:
\begin{center}
\texttt{WFMT$_0$|WFMT$_1$|...[\^{}N][sSFMT][+PFMT][.OPQ$\sigma$][.Wr$\rho$]}
\end{center}
where each bracketed component is optional.

\paragraph{WFMT (weight format).} The per-block LUT alphabet.
Multiple WFMTs separated by \texttt{|} indicate per-block
selection via a metadata bit, e.g. \textsf{NF4|DD4} chooses
between the fixed NF4 codebook (LUT0) and a per-layer-trained
DD4 codebook (LUT1) per block.  Single-WFMT configs omit the
pipe.  Atoms used in this paper: \textsf{E2M3} (FP6),
\textsf{E4M3} (FP8), \textsf{NF4} / \textsf{NF5}
(NormalFloat), \textsf{DD4} / \textsf{DD5} (per-layer
Dynamic Decode), \textsf{Split87}, \textsf{SH4} / \textsf{SH5},
\textsf{MPO2}, \textsf{BOF4}.

\paragraph{\^{}N (block size).} The per-block scope.  \^{}16
is default and may be elided.  \^{}0 indicates layer-max
scope (one scale per layer).  Block sizes 8 and 32 are
admissible but not used in this paper's experiments.

\paragraph{sSFMT (scale format).} The per-block scale format
in container size $w{+}x{+}y$ bits.  \textsf{UE}$x$\textsf{M}$y$
is unsigned (no sign bit); \textsf{E}$x$\textsf{M}$y$ is signed
(sign bit may double as a metabit for atom-pair selection or
per-block polarity).  \textsf{UE8M0} is power-of-two only;
mantissa-bearing variants (e.g. \textsf{UE4M3}, \textsf{UE4M6},
\textsf{UE5M3}) carry fractional precision.

\paragraph{+PFMT (promotion).} The methodology promotes
worst-quantized layers.  A fixed target \texttt{+}PFMT
promotes selected layers to PFMT (common values:
\textsf{E2M3sUE4M3} at 6.5 bpw, \textsf{E4M3\^{}} at 8.0 bpw
layer-max, \textsf{E5M6\^{}} at 12.0 bpw, \textsf{E8M7} at
16.0 bpw BF16).  Alternatively, \texttt{+knap}$N$ invokes the
MCKP allocator (\S\ref{sec:mckp}) at a budget of $+N/10$ bpw
above base, jointly selecting layers and PFMTs; an optional
\texttt{/PFMT/} suffix list restricts the candidate set.

\paragraph{.OPQ$\sigma$ (outlier per-quantum extraction).}
Sparse weight-outlier extraction at sigma-threshold $\sigma$
(\S\ref{sec:opq}).

\paragraph{.Wr$\rho$ (sparse residual correction).}
Activation-weighted sparse residual correction at sparsity
$\rho$ percent (\S\ref{sec:wr}).  Bare \textsf{.Wr} defaults
to $\rho = 0.1\%$.

\paragraph{Examples.}
\begin{itemize}\itemsep0pt
\item \textsf{E2M3sUE4M4} --- FP6 weight, per-block (g=16)
      unsigned UE4M4 scale.  $6.5$ bpw.
\item \textsf{E4M3\^{}0sUE8M0} --- FP8 weight, layer-max
      power-of-two scale.  $8.0$ bpw.
\item \textsf{NF4|DD4sUE4M3+knap0.10.Wr} --- NF4/DD4 pair,
      per-block UE4M3 scale, knapsack promotion at $+0.10$
      bpw budget, default-sparsity Wr correction.  $\sim 4.7$
      bpw effective.
\end{itemize}

\end{document}